\documentclass[letterpaper]{article} 
\usepackage{aaai25}  
\usepackage{times}  
\usepackage{helvet}  
\usepackage{courier}  
\usepackage[hyphens]{url}  
\usepackage{graphicx} 
\usepackage{natbib}  
\usepackage{amsmath}
\usepackage{mathtools}
\usepackage{float}
\usepackage{tabularx}
\usepackage{caption} 
\usepackage{algorithm}
\usepackage{algorithmic}
\usepackage{xcolor}
\usepackage{amsfonts}
\usepackage{soul}
\usepackage{amssymb}
\usepackage{booktabs}      
\usepackage{multirow}
\usepackage{todonotes}
\usepackage{threeparttable}
\usepackage{svg}
\usepackage{stackengine}
\usepackage{bm}
\usepackage[shortlabels]{enumitem}
\usepackage{footmisc}
\usepackage{tablefootnote}
\usepackage{newfloat}
\usepackage{listings}
\usepackage{amssymb}
\usepackage{pifont}
\usepackage{stackengine}
\newcommand{\cmark}{\ding{51}}%
\newcommand{\xmark}{\ding{55}}%
\newcommand{\vect}[1]{\mathbf{\bm{#1}}}

\DeclareCaptionStyle{ruled}{labelfont=normalfont,labelsep=colon,strut=off} 
\floatstyle{ruled}
\newfloat{listing}{tb}{lst}{}
\floatname{listing}{Listing}

\title{PNVC: Towards Practical INR-based Video Compression}

\author{
    Ge Gao, 
    Ho Man Kwan,
    Fan Zhang,
    David Bull
}
\affiliations{
    Bristol Visual Institute, University of Bristol\\
    \{ge1.gao, hm.kwan, fan.zhang, dave.bull\}@bristol.ac.uk   
}

\usepackage{bibentry}

\begin{document}

\maketitle

\begin{abstract} \label{sec:abstract}
Neural video compression has recently demonstrated significant potential to compete with conventional video codecs in terms of rate-quality performance. These learned video codecs are however associated with various issues related to decoding complexity (for autoencoder-based methods) and/or system delays (for implicit neural representation (INR) based models), which currently prevent them from being deployed in practical applications. In this paper, targeting a practical neural video codec, we propose a novel INR-based coding framework, PNVC, which innovatively combines autoencoder-based and overfitted solutions. Our approach benefits from several design innovations, including a new structural reparameterization-based architecture, hierarchical quality control, modulation-based entropy modeling, and scale-aware positional embedding. Supporting both low delay (LD) and random access (RA) configurations, PNVC outperforms existing INR-based codecs, achieving nearly 35\%+ BD-rate savings against HEVC HM 18.0 (LD) - almost 10\% more compared to one of the state-of-the-art INR-based codecs, HiNeRV and 5\% more over VTM 20.0 (LD), while maintaining 20+ FPS decoding speeds for 1080p content. This represents an important step forward for INR-based video coding, moving it towards practical deployment. The source code will be available for public evaluation.
\end{abstract}

%

\section{Introduction} 
There is an ever-increasing requirement for higher video coding efficiency due to the significantly increased demand for high quality digital video content. Unlike standardized video codecs, such as H.265/HEVC~\cite{sullivan2012overview} and H.266/VVC~\cite{bross2021overview} that offer impressive performance using evolutions of conventional architectures, neural video compression is enjoying much faster development cycles and rapidly increasing performance benchmarks using an optimized, data-driven end-to-end architecture. Advances in this research area have delivered a wide variety of candidate neural video codecs, some of which~\cite{xiang2022mimt,li2024neural} are reported to match or outperform the (rate-quality) performance of the latest state-of-the-art standard coding methods.

  \begin{figure}[!t]
    \centering
    \includegraphics[width=\linewidth]{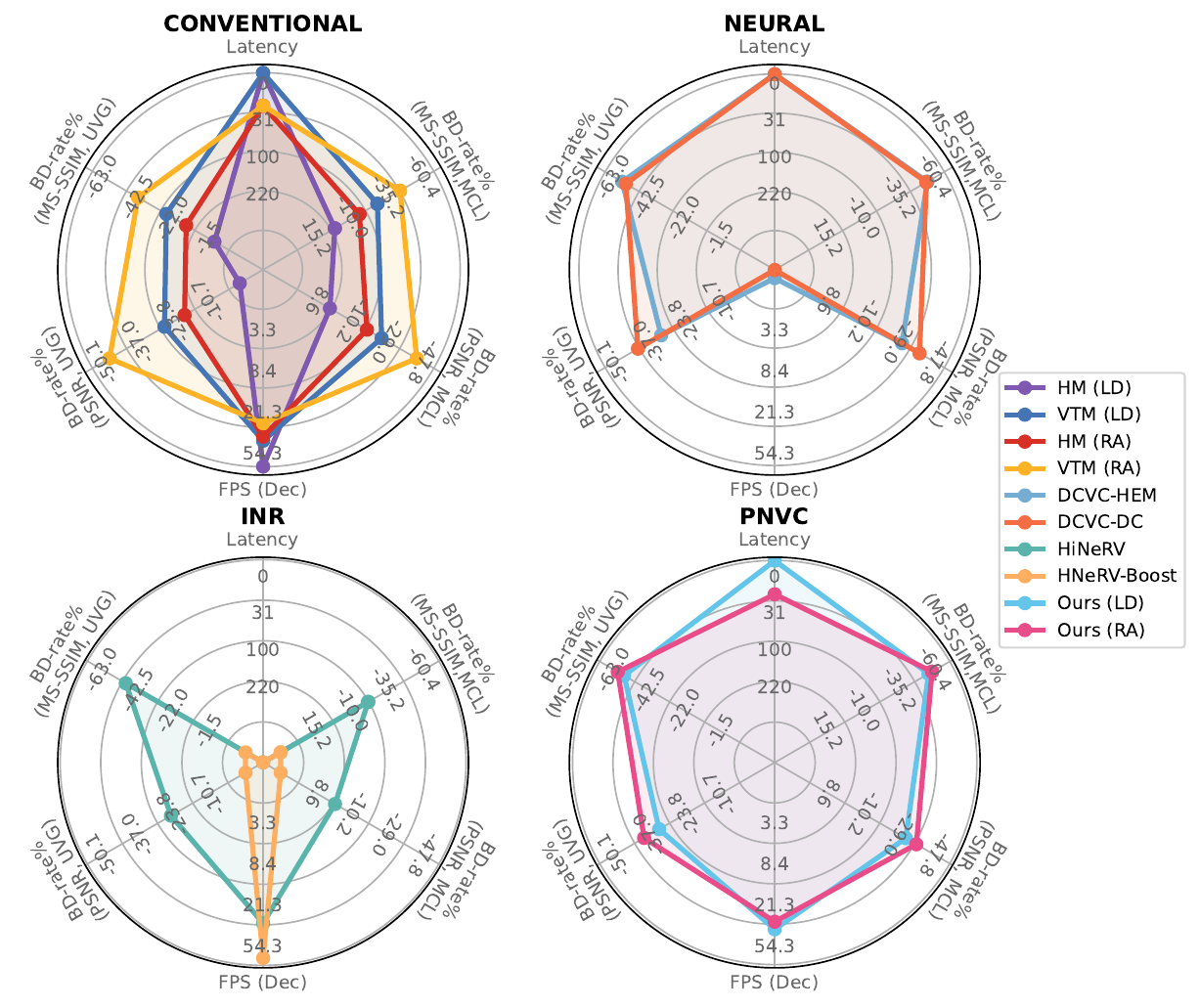}
    \caption{Radar plots illustrating the performance of proposed PNVC codec (ours) and nine other conventional and neural video codecs, in terms of coding efficiency (BD-rate measured by PSNR and MS-SSIM on UVG and MCL-JVC datasets, against HM 18.0, LD),  decoding speeds (FPS) and coding latency\textsuperscript{\ref{fn:latency}} (frames). It can be observed that PNVC demonstrates excellent performance in all these aspects.
    }
    \label{fig:performance-complexity} 
\end{figure}

Although showing promise in terms of coding gain, neural video codecs (primarily those using autoencoder-based backbones) are associated with significant complexity issues, in particular on the decoder side, making them resource-intensive and impractical for many real-world applications. Although common complexity reduction techniques such as pruning and quantization~\cite{peng2023accelerating,hu2023complexity} can alleviate these limitations, this typically results in a nontrivial reduction in coding efficiency. 

More recently, implicit neural representation (INR)-based coding methods~\cite{chen2021nerv,he2023towards,kwan2024hinerv} have attracted increasing attention as a paradigm-shifting solution to achieving high coding performance and low (decoding) complexity. This type of approach typically utilizes a lightweight neural network to overfit input video data by mapping coordinates directly to pixel values. Although the latest INR-based codecs~\cite{kwan2024hinerv} have shown consistent coding gains over many conventional and neural video codecs, they have a major limitation in that their compression strategy to represent an entire video sequence or a dateset with a single monolithic model. While this practice maximizes compression efficiency, it requires processing a large number of video frames  (e.g., a few hundred to a few thousand) in each encoding session, which conflicts with commonly used coding configurations, where flexible system latency\footnote{\label{fn:latency}Here latency is defined as the system delays in the coding process, e.g., the number of consecutive bi-directional predicted frames in a GOP (Group of Pictures).}, e.g., Low Delay and Random Access modes in VVC VTM~\cite{vtm_ctc}, is often required. This issue prevents INR-based video codecs from being as performant on shorter sequences or adopted in many practical applications.

In this paper, we introduce PNVC, a novel (\textbf{P}ractical) I\textbf{N}R-based \textbf{V}ideo \textbf{C}ompression framework that tackles the aforementioned limitations, which enables flexible coding configurations (low latency) while still achieving competitive coding performance and low encoding/decoding complexity. The proposed PNVC is built upon a hierarchical backbone that generalizes autoregressive models~\cite{lu2024deep} and is seamlessly interchangeable with content- or modulation-based INR models~\cite{lee2024coordinate,kwan2024hinerv}. Our approach leverages a \textit{pretrain-then-overfit} strategy, enabling the model to generalize across diverse content during pretraining whilst adapting to input-specific contents during overfitting. Further, we develop a novel reparameterization method, among other architectural and optimization innovations, that allows for unconstrained model capacity during training while ensuring low-complexity inference. This decoupling enables more effective optimization without sacrificing efficiency in deployment. The main contributions of this paper are summarized as follows.

\begin{itemize}
    \item [1)] We propose a \textbf{novel INR-based video coding model} that integrates autoencoder-based with overfitted solutions, offering competitive coding performance, relatively low encoding and decoding complexity and a flexible coding latency configuration.
    
    \item [2)] We design a novel \textbf{reparameterization}-based scheme (ModMixer) to sufficiently pretrain as well as overfitting a lightweight backbone with stronger modeling capacity and more diverse optimization directions, without extra inference costs. 

    \item [3)] We further introduce several modifications, including \textbf{hierarchical quality parameters}, \textbf{modulation-based hierarchical entropy model} with asymmetric context grouping, and \textbf{scale-aware hierarchical positional encoding} to enhance the compression performance.
\end{itemize}

The proposed PNVC demonstrates very competitive rate-distortion performance in both Low Delay and Random Access configurations (as defined in many video coding standards), whilst circumventing the latency and encoding complexity problems associated with existing INR-based video codecs. Specifically, as shown in Figure \ref{fig:performance-complexity}, on both UVG and MCL-JCV, the proposed model evidently outperforms VTM 20.0 (LD), and HiNeRV in BD-rate, measured in both PSNR and MS-SSIM. It is also associated with much lower encoding latency (system delays) compared to existing INR-based videos, and a faster decoding speed over autoencoder-based neural video coding models. We believe that this work yields a major step forward in INR-based video compression, moving it toward practical adoption.

\section{Related Work}

\subsubsection{Video compression} has been a long-standing research topic, evolving from hand-crafted standard codecs~\cite{wiegand2003overview,sullivan2012overview,bross2021overview} to deep learning techniques that enhance conventional tools~\cite{yan2018convolutional,ma2020mfrnet,zhang2021video} and end-to-end optimized frameworks~\cite{lu2019dvc,agustsson2020scale}. Recent learning-based methods have focused on improving motion estimation~\cite{li2023neural}, feature space conditional coding~\cite{hu2021fvc,li2021deep}, instance-adaptive overfitting~\cite{van2021overfitting,khani2021efficient,yang2024parameter}, and novel architectures~\cite{ho2022canf,xiang2022mimt,mentzer2022vct}. Despite competitive coding performance, high computational complexity (in particular in the decoder) has limited practical deployment, with structured pruning approaches~\cite{peng2023accelerating} achieving only a limited reduction in complexity at the cost of significant performance drops.

\subsubsection{Implicit Neural Representations} (INRs) have been increasingly employed in recent years to represent and compress various multimedia signals, including images~\cite{dupont2021coin,strumpler2022implicit}, videos~\cite{zhang2021implicit,kwan2024hinerv} and volumetric content~\cite{ruan2024point,kwan2024immersive}. INRs learn a coordinate-based mapping function and encode data in network parameters. Existing implicit neural video representation (NeRV) models can be categorized as: i) \textit{index-based} methods taking frame~\cite{chen2021nerv}, patch~\cite{bai2023ps}, or disentangled spatial/grid coordinates~\cite{li2022nerv} as input, or ii) \textit{content-based} methods~\cite{chen2023hnerv,kwan2024hinerv,kim2024c3,leguay2024cool} with content-specific embeddings as inputs. In these cases, the video coding task is reformulated into a model compression problem, leveraging pruning, quantization, and entropy-constraint optimization~\cite{gomes2023video}. However, training NeRV models on entire video sequences or datasets leads to high system latency\footref{fn:latency}, making them unsuitable for applications requiring quick response times, and leading to less meaningful comparisons with latency-constrained codecs.

\begin{figure*}[!hbt]
    \centering
    \includegraphics[width=0.95\linewidth]{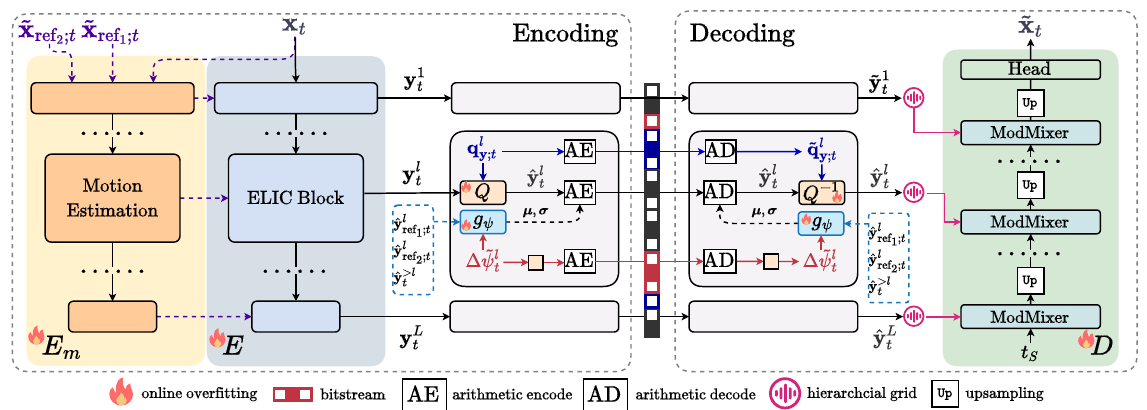}
    \caption{Illustration of the proposed PNVC framework.}
    \label{fig:network}
\end{figure*}

\section{Methods} \label{sec:methods}

\subsection{Overview}
In the proposed PNVC framework, each video sequence $\{ \vect{x}_t \}^{T}_{t=1},\ \vect{x}_t \in \mathbb{R}^{3 \times H \times W} $, is segmented into Groups of Pictures (GOPs)\footnote{We follow the GoP definition in H.266/Versatile Video Coding Test Model (VTM), where the length of a GoP equals the number of consecutive bi-directional predicted (B) frames plus 1.} of length $N$ that are independently encoded, where $T$, $H$, and $W$ represent the length, height, and width of $\{ \vect{x}_t \}^{T}_{t=1}$, respectively. Within each GOP, frames are either intra-coded as I frames or inter-coded as P/B frames to exploit spatial and/or temporal redundancy within the video.

Each I frame is encoded into $L$ grids of latent tokens $\{\vect{y}_t^{l}\}^{L}_{l=1}$ by an encoder $E$. For each level $l$, $\vect{y}_t^{l}$ consists of rescale and shift parameters $\vect{\gamma}^{l}_t$ and $\vect{\beta}^{l}_t$, concatenated channel-wise, with a resolution of $\left(\frac{H}{2^l}, \frac{W}{2^{l}}\right)$.
A learnable quantizer $Q$ then quantizes these latent grids, producing $\{\hat{\vect{y}}_t^l\}^{L}_{l=1}=\{ \hat{\vect{\gamma}}^{l}_t \| \hat{\vect{\beta}}^{l}_t \}^{L}_{l=1}$ based on the corresponding quantization parameters $\{\vect{q}^{l}_{\vect{y};t}\}^L_{l=1}$, where $\|$ denotes the channel-wise concatenation. 

An entropy model $g_{\vect{\psi}}$ is adopted to estimate the probability mass function (PMF) of the hierarchical latents semi-autoregressively along the spatial, channel, and hierarchical axes. The parameter of the entropy model, $\vect{\psi}$, is updated as $\vect{\psi}^{l}_t \rightarrow \vect{\psi} + \Delta \tilde{\vect{\psi}}^l_t$ at each level $l$. Here, $\vect{\psi}$ represents the pretrained parameters, and $\Delta \tilde{\vect{\psi}}^{l}_t$ are dequantized parameter updates obtained through online overfitting. The bitstream includes $\{ \hat{\vect{y}}^l_t, \Delta \hat{\vect{\psi}}^{l}_t, \tilde{\vect{q}}^{l}_{\vect{y};t}, \tilde{q}^{l}_{\vect{\psi};t}\}^L_{l=1}$, where $\{ \tilde{\vect{q}}^{l}_{\vect{y};t} \}^L_{l=1}$ and $\{\tilde{q}^{l}_{\vect{\psi};t}\}^L_{l=1}$ are the inverse quantization parameters corresponding to latents and weight updates, respectively. These components are overfitted to the input video frame, entropy encoded, and combined into the bitstream.

At the decoder, the entropy model is updated by $\{ \tilde{\vect{\psi}}^{l}_t \}^L_{l=1}$ decoded from the bitstream. It is used to help entropy decode the hierarchical latent grids that are then dequantized by $Q^{-1}$ using $\{\tilde{\vect{q}}^{l}_{\vect{y};t}\}^L_{l=1}$ to produce $\{\tilde{\vect{y}}^{l}_t\}^L_{l=1}$. These dequantized latents undergo positional embedding based on an improved scale-aware hierarchical decomposition scheme from~\cite{kwan2024hinerv}, allowing content-specific variations to be queried by spatial-temporal coordinates. A patch-based representation is used, with each element of the smallest $\tilde{\vect{y}}^{l}_t$ corresponding to a patch of the original frame $\vect{x}_t$. The INR-based decoder $D$ defines a patch-wise neural field as a function of 3D coordinates. It uses $L$ stacked ModMixer blocks (proposed in this work) to progressively and conditionally map $t_S$, which is a learned bias for I-frame and the reference patch for P-/B- frames, to patches of the reconstructed frame $\tilde{\vect{x}}_t$, with the intermediate activations of each block modulated by the corresponding latent grid.

Each P or B frame is coded following a workflow similar to that of I frames, but with an additional motion encoder $E_m$, as shown in Figure~\ref{fig:network}. With up to two previously reconstructed frames, $\tilde{\vect{x}}_{\text{ref}_1;t}$ and $\tilde{\vect{x}}_{\text{ref}_2;t}$, in the decoded frame buffer (they can be I or P/B frames) and the current frame $\vect{x}_t$, the multi-resolution latent grids $\{ \vect{y}_t^{l} \}^{L}_{l=1}$ are generated by $E$ but with layer-wise conditioning on the motion information extracted from $E_\text{m}$. Here, the entropy model $g_{\vect{\psi}}$ additionally takes into account the decoded reference latent tokens $\tilde{\vect{y}}^{l}_{\text{ref}_1;t}, \tilde{\vect{y}}^{l}_{\text{ref}_2;t}$ and the reconstructed latent tokens from higher-level layers, $\tilde{\vect{y}}^{>l}_t$, as input to further exploit spatio-temporal and hierarchical redundancy.

\subsection{Encoder}
The ELIC ~\cite{he2022elic} model is adopted here as the image encoder $E$ to map I/P/B frames to the latent $\{\vect{y}^{l}_t\}^L_{l=1}$. For P/B frames, $E$ is reconditioned by concatenating the hierarchical optical flow features, extracted by the motion encoder $E_m$, adapted from SpyNet~\cite{ranjan2017optical}, to incorporate the estimated motion information. The illustration of its network structure is provided in the \textit{Supplementary}. Both $E$ and $E_m$ are overfitted to generate content-adaptive hierarchical latent representations. The overfitting of the next frame is initialized from the network updated for the current frame to further expedite the encoding process.

\subsection{Quantization} 
A hierarchical quality structure~\cite{li2024neural} is adopted in the quantization module, where the allowable bitrates are adaptively reweighted for each frame based on its distance from reference frames and specific video dynamics. This parameter is referred to as the quality parameter $q_{\text{glob};t}$, generated by a ConvLSTM module based on each token's estimated impact on the GOP-level RD trade-offs. This module is fixed after pretraining to prevent the frame-wise overfitting process from destroying the acquired hierarchical quality structure. For quantization, another set of finer-grained channel-wise quality parameters $\{\vect{q}^{l}_{\vect{y};t}\}^{L}_{l=1}$ are used. For inverse quantization, the corresponding $\{\tilde{\vect{q}}^{l}_{\vect{y};t}\}^L_{l=1}$ are retrieved and encoded into the bitstream as side information. The quantization and inverse quantization steps are exemplified using $\vect{y}^l_t$ as:
\begin{subequations}
    \begin{align}
        \hat{\vect{y}}^{l}_t[c][i][j] &= \lfloor \vect{q}^{l}_{\vect{y};t}[c] \cdot \vect{y}^{l}_t \rceil, \\
        \tilde{\vect{y}}^{l}_t[c][i][j] &= \tilde{\vect{q}}^{l}_{\vect{y};t}[c] \cdot \hat{\vect{y}}^{l}_t,
\end{align}
\end{subequations}
where \(c\), \(i\), and \(j\) denote the channel and spatial indices, respectively. Here, we avoid division in both cases to avoid numerical instabilities. The scalar quantization parameters $\{ q_{\vect{\psi}}\}^L_{l=1}$ are updated following a similar procedure.

\subsection{Entropy coding}
The discretized hierarchical latents $\{\hat{\vect{y}}^{l}_t\}^L_{l=1}$ are entropy coded based on arithmetic coding. Here, PMFs are estimated by a Gaussian distribution $\mathcal{N}(\cdot)$, where the location and scale parameters $\left( \vect{\mu}_{t}[c][i][j], \vect{\sigma}_{t}[c][i][j] \right)$ of each element $\hat{\vect{y}}^l_{t}[c][i][j]$ are semi-autoregressively predicted by the entropy network $g_{\vect{\psi}}$ based on the spatial, temporal, and hierarchical contexts of the element. We use the quadtree-based factorization~\cite{li2023neural} to encode $\hat{\vect{y}}^{l}_t$, but with two modifications. First, we respectively split the channels of $\hat{\vect{\gamma}}^l_t$ and $\hat{\vect{\beta}}^l_t$ (they are entropy coded in parallel) into four uneven groups~\cite{he2022elic}, with sizes proportional to 1, 1, 2 and 4, \textit{i.e.} we double the number of symbols decoded per decoding step due to more contexts available. Moreover, we replace all concatenation operations to aggregate information with element-wise modulation, which we empirically found to be far more efficient in information aggregation, and deploy a new ModMixer module (detailed in the next subsection) to construct and optimize the entropy model. The architecture illustration of the entropy model is provided in the \textit{Supplementary}. For entropy model's updates $\{\Delta \hat{\vect{\psi}}_t^{l}\}_{l=1}^{L}$, a fully factorized nonparametric density model $\vect{\pi}$~\cite{balle2018variational} is deployed following~\cite{gomes2023video,zhang2024boosting}.

\subsection{ModMixer} 

\begin{figure}[!t]
    \centering
    \includegraphics[width=0.85\linewidth]{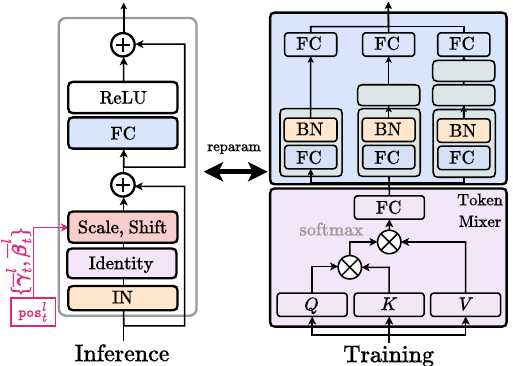}
    \caption{The architectures of the reparamterized ModMixer block during training and inference.}
    \label{fig:reparam}
\end{figure}

To enhance model performance without introducing extra inference cost, we devise a novel \textbf{ModMixer} module as a basic building block used in our entropy model $g_{\vect{\psi}}$ and decoder $D$. This approach is inspired by reparameterization methods~\cite{ding2021repmlp,ding2022scaling,shi2024improved} that leverage the interconversion between linear architectures to trade training-time complexity for inference-time efficiency. Without loss of generality, we define an arbitrary linear layer (convolutional or fully-connected), $\vect{W} \vect{h} + \vect{b}$, to be contracted algebraically from this general formulation:
\begin{equation}
 \sum_{m=1}^M \left( \overline{\vect{\Lambda}}_{m;\vect{\gamma_W}} \vect{W}_m + \overline{\vect{\Lambda}}_{m;\vect{\beta_W}} \right) \vect{h} + \left(\overline{\vect{\Lambda}}_{m;\vect{\gamma_b}} \vect{b}_m + \overline{\vect{\Lambda}}_{m;\vect{\beta_\vect{b}}} \right),  \label{eq:reparam}
\end{equation}
where $\vect{W}_m \in \mathbb{R}^{c_{\text{out}}\times c_{\text{in}}}$ and $\vect{b}_m \in \mathbb{R}^{c_\text{out}}$ denote the trainable, parallel basis of $\vect{W}$ and $\vect{b}$, respectively, and the coefficients $\{ \overline{\vect{\Lambda}}_{m;\vect{\gamma_W}}, \overline{\vect{\Lambda}}_{m;\vect{\beta_W}}, \overline{\vect{\Lambda}}_{m;\vect{\gamma_b}}, \overline{\vect{\Lambda}}_{m;\vect{\beta_b}} \} \in \mathbb{R}^{c_{\text{out}}}$ denote the channel-wise affine mixand values. Here, each basis may be further expanded serially into (taking $\vect{W}_m$ as an example) $\vect{W}_m = \vect{W}_{m;N} \vect{W}_{m;N-1} \cdots \vect{W}_{m;1}$. Compared to standard reparameterization methods, Equation~(\ref{eq:reparam}) establishes a more generalized visual tuning scheme that better suits the context of instance-adaptive video compression. With the affine transformation applied to both weight and bias terms, it is a \textit{superset} encompassing both weight and feature space tuning, where the mixands could be flexibly re-casted to feature modulation (as in decoder), weight update (as in entropy model), or any other forms of visual tuning accordingly. 

In our approach, to leverage the above reparameterization idea and improve fully connected layers (FC), typically adopted in INR-based methods~\cite{sitzmann2020implicit,dupont2021coin}, a new basic building block, ModMixer, is designed as illustrated in Figure~\ref{fig:reparam}. Each FC layer is reparameterized both serially and in parallel through stacks of linear layers with varying depths per branch (one per channel group) to capture the hierarchical representations efficiently. Further, a self-attention-style token-mixer is attached before the FC layer to induce inference-time spatial mixing. During pretraining, the token mixer is progressively \textit{degenerated} and absorbed into the Instance Normalization~\cite{huang2017arbitrary} and proceeding FC layer~\cite{lin2024mlp}, by initializing a mask $\vect{M}$ as $\vect{1}$ that is gradually decayed to $\vect{0}$: $\vect{M} \odot \text{TokenMixer}\left( \vect{h} \right) + \left(1 - \vect{M} \right) \odot \vect{h} + \vect{h}$.

The mask decaying strategy is inspired by~\cite{wang2023EVC,peng2023accelerating}, which leverage the more expressive teacher model to distill the smaller student. During overfitting, we optimize the residual updates, w.r.t the pretrained mixands and basis, which are then consolidated into either weight updates $\Delta \vect{\psi}^l_t$ for the entropy model or absorbed into the hierarchical latent grids $\vect{y}_t^l$ for the decoder (see \textit{Supplementary} for full derivations and implementation details).

\subsection{Decoder}
In Figure~\ref{fig:network}, the decoder $D(\cdot)$ acting on the grid-based positional encodings incrementally restores spatial information while sequentially composing high-frequency details over low-frequency elements. We follow the hierarchical encoding method proposed by the high-performance HiNeRV~\cite{kwan2024hinerv}, which can be conceptualized as a positional numeral system with power-of-2 bases. This method expresses each coordinate as an ordered set of digits, each of which corresponds to a hierarchy that recursively encodes residuals of the coarser-grained representational level. Specifically, this hierarchical coordinate system decomposes a global coordinate $\texttt{pos}$ into multiple levels of finer detail with base $B_l$. At each level $l$, the local coordinate for interpolation is computed as:
\begin{equation}
    \texttt{pos}^l = \left\lfloor \frac{\texttt{pos}}{\prod_{k=L}^{l-1} B_k} \right\rfloor \mathrm{\ mod\ } B_l.
\end{equation}

We make a simple improvement to the original strategy to cope with the increase in patch size at inference time compared to that seen by the model during pretraining. The likely substantial increase in resolution, denoted by $k$, could result in more fine-tuning endeavours. Maintaining the original grid spacing would result in an increase in overall grid numbers, potentially complicating hierarchical pattern capturing and increasing memory/computational demands To address this, we propose to instead \textit{mix and rescale} the bases $\{ B_l \}^L_{l=1}$ non-linearly according to the ratio ($k$). The local coordinate at level $l$ is re-calculated as:
\begin{equation}
    \texttt{pos}^l = \left\lfloor \frac{\texttt{pos}}{\prod^{l-1}_{k=L}  w_k B_k} \right\rfloor \mathrm{\ mod\ } (w_l B_l), \label{eq:scaled-pe}
\end{equation}
subject to $w_1w_2\cdots w_L = k$ and $w_1 \geq w_2 \geq \cdots \geq w_L \geq 1$. Solving this constraint yields a set of $\{ w_l \}^L_{l=1}$ evenly distributed on a log-scale, as will be detailed in \textit{Supplementary}. The formulation by Equation~(\ref{eq:scaled-pe}) essentially distributes the interpolation pressure across different hierarchies. Low frequencies, which are typically less sensitive to resolution changes, can handle more scaling without significant loss of information (sparser grid partition at lower resolution). High frequencies, on the other hand, are preserved more carefully from being scaled less. Based on Equation (\ref{eq:scaled-pe}), the $l$-th layer of decoder is given by:
\begin{gather*}
    \ddot{\vect{\gamma}}^{l}_t, \ddot{\vect{\beta}}^{l}_t = \text{Interp}_{\text{bilinear}} \left(\tilde{\vect{\gamma}}^{l}_t, \texttt{pos}^{l}\right), \text{Interp}_{\text{blinear}}\left(\tilde{\vect{\beta}}^{l}_t,\texttt{pos}^{l}\right), \\
    \ddot{\vect{h}}_t^{l-1} = \text{ReLU} \left(\vect{W}^{l} \left( \left( \ddot{\vect{\gamma}}^{l}_t \text{IN}\left(\vect{h}^{l}_t\right) + \ddot{\vect{\beta}}^{l}_t \right) + \vect{h}^l_t \right) + \vect{b}^{l} \right), \\
    \vect{h}_t^{l-1} = \text{Upsample}_{\text{bilinear}}(\ddot{\vect{h}}^{l-1}_t; S^{l-1}_{\text{default}}=2).
\end{gather*}

\subsection{Optimization strategy}
The full model is first pretrained offline, based on static training materials, and then overfitted during inference to adapt to the input video sequence to be compressed. Considering the pre-trained parameters, the ``meta-initialization'' shared between the encoder and the decoder, the iteratively refined parameter updates, relative to this meta-initialization, are quantized alongside associated side information and then entropy coded into the bitstream. We follow~\cite{lu2020content,li2023neural} to aggregate the loss over multiple frames to reduce error propagation and establish the hierarchical quality structure. The pretraining involves minimizing the following rate-distortion loss within each GoP:
\begin{equation}
    \frac{1}{N} \sum^N_{t=1} \left( \sum^{L}_{l=1} \left(R(\hat{\vect{y}}^{l}_t) + R(\hat{\vect{q}}^{l}_{\text{ch};t})\right) + q_{\text{glob};t} \cdot \lambda \cdot d(\vect{x}_t, \hat{\vect{x}}_t) \right) , \label{eq:rd-prime}
\end{equation}
where $d(\cdot, \cdot)$ denotes the distortion, $R(\cdot)$ stands for the rate, and $t \in {1, \cdots, N}$ denotes the displaying order of each frame in a GOP. Following~\cite{kim2024c3,leguay2024cool}, the quantization of hierarchical latents and weight update is approximated by progressively annealed soft-rounding with additive noises when calculating $R(\cdot)$, and by the straight-through estimator (STE)~\cite{minnen2020channel} when rounding them to optimize the distortion metric. Here MSE is used as the distortion metric targeting the best PSNR performance, while additional models are trained by fine-turning MSE models using $200 \cdot (1 - \text{MS-SSIM}(\vect{x}_t, \hat{\vect{x}}_t))$~\cite{mentzer2022vct} as the distortion metric to yield MS-SSIM-based baselines. During frame-wise overfitting, encoding entails searching for the optimal values of the hierarchical latents, network parameters, and quantization parameters:
\begin{equation}
     \sum^{L}_{l=1} \left(R(\hat{\vect{y}}^{l}_t)+R(\Delta \hat{\vect{\psi}}^{l}_t) + R(\hat{\vect{q}}^{l}_{\text{ch};t}) \right) + q_{\text{glob};t} \cdot \lambda \cdot d(\vect{x}_t, \hat{\vect{x}}_t).\label{eq:rd}
\end{equation}
Detailed hyperparameter configurations could be found in the \textit{Supplementary}.

\begin{table*}[htb!]
    \caption{BD-rate results w.r.t H.265/HEVC Test Model HM 18.0 (LD), denoted by *, together with the complexity figures including kMACs/pixel, FPS, and model size, on the UVG dataset. Complexity figures for INR benchmarks are reported with $\pm$ to indicate the range as their sizes and configurations change for different bitrates.}
    \centering
    \label{tab:bd-psnr}
    \resizebox{\textwidth}{!}
    {
    \begin{tabular}{r|rr|rr|cc|c|c|c}
        \toprule
         & \multicolumn{2}{c|}{UVG Dataset} & \multicolumn{2}{c|}{MCL-JCV Dataset} & \multicolumn{4}{c|}{Model Complexity (UVG)}& \\
        \cmidrule{2-9}
        Codec (mode) &  \multicolumn{2}{c|}{BD-rate} & \multicolumn{2}{c|}{BD-rate} & \multicolumn{2}{c|}{FPS} & \multicolumn{1}{c|}{Params. (M)} & \multicolumn{1}{c|}{kMACs/pixel} & Latency\\ 
        \cmidrule{2-9}
        & (PSNR) & (MS-SSIM) & (PSNR) & (MS-SSIM) & Encoding & Decoding & Full model & Decoding \\
        \midrule
        *HM 18.0 (LD) & -0.00\% & -0.00\% & -0.00\% & -0.00\%  & 6.17 & 54.3 & N/A & N/A & 0 \\
        HM 18.0 (RA) & -21.21\% & -17.02\% & -20.36\% & -18.10\% & 3.48 & 33.2 & N/A & N/A & 31 \\
        VTM 20.0 (LD) & -28.94\% & -28.90\% & -28.47\% & -31.35\% & 0.05 & 30.7 & N/A & N/A & 0 \\
        VTM 20.0 (RA) & -50.08\% & -45.94\% & -47.78\% & -48.02\% & 0.001 & 23.8 & N/A & N/A & 31 \\
        \midrule
        VCT & +20.12\% & -42.06\% & +20.70\% & -48.57\% & 1.10 & 0.39 & 154.2 & 5336 & 0 \\
        DCVC-HEM & -34.96\% & -63.05\% & -33.81\% & -60.35\% & 0.87 & 1.51 & 48.7 & 1581 & 0 \\
        DCVC-DC & -43.94\% & -60.83\% & -43.46\% & -59.65\% & 0.96 & 1.29 & 50.8 & 1274 & 0 \\
        \midrule
        FFNeRV & +81.50\% & +50.21\% & +100.93\% & +80.31\% & $\text{0.0336} \pm \text{0.013}$ & $\text{36.4} \pm \text{2.7}$ & $\text{12.1} \pm \text{8.7}$ & $\text{36.8} \pm \text{1.6}$ 
        & Full seq.\\
        C3 (Adaptive)\tablefootnote{For C3, we adopt the adaptive setting for the UVG dataset, where the patch length ranges from 30 to 75 for different rates yet is fixed to 25 on the MCL-JCV dataset.} & -0.75\% & -15.53\% & +20.64\% & +40.76\% & 0.0015 & 17.6 & 0.01 & 4.4  
        & 25-75 \\
        HNeRV-Boost & +2.42\% & +18.96\% & +75.8\% & +225.6\% & $\text{0.0155} \pm \text{0.009}$ & $\text{47.6} \pm \text{24.3}$ & $\text{11.6} \pm \text{8.9}$ & $\text{520.0} \pm \text{359.0}$ & Full seq. \\
        HiNeRV & -25.42\% & -50.90\% & -1.64\% & -22.34\% & $\text{0.0157} \pm \text{0.009}$ & $\text{19.7} \pm \text{13.5}$ & $\text{30.3} \pm \text{27.1}$ & $\text{682.8} \pm \text{595.1}$ & Full seq. \\
        \midrule
        \textbf{PNVC (LD, ours)} & \textbf{-34.35\%} & \textbf{-58.97\%} & \textbf{-33.56\%} & \textbf{-57.73\%} & 0.014 & 25.3 & 21.8 & 101.1 & 0 \\
        \textbf{PNVC (RA, ours)} & \textbf{-39.98\%} & \textbf{-62.72\%} & \textbf{-39.23\%} & \textbf{-60.21\%} & 0.011 & 22.6 & 21.8 & 101.1 & 31 \\
        \bottomrule
    \end{tabular}}
\end{table*}

\begin{figure*}[htb!]
    \centering
    \includegraphics[width=\linewidth]{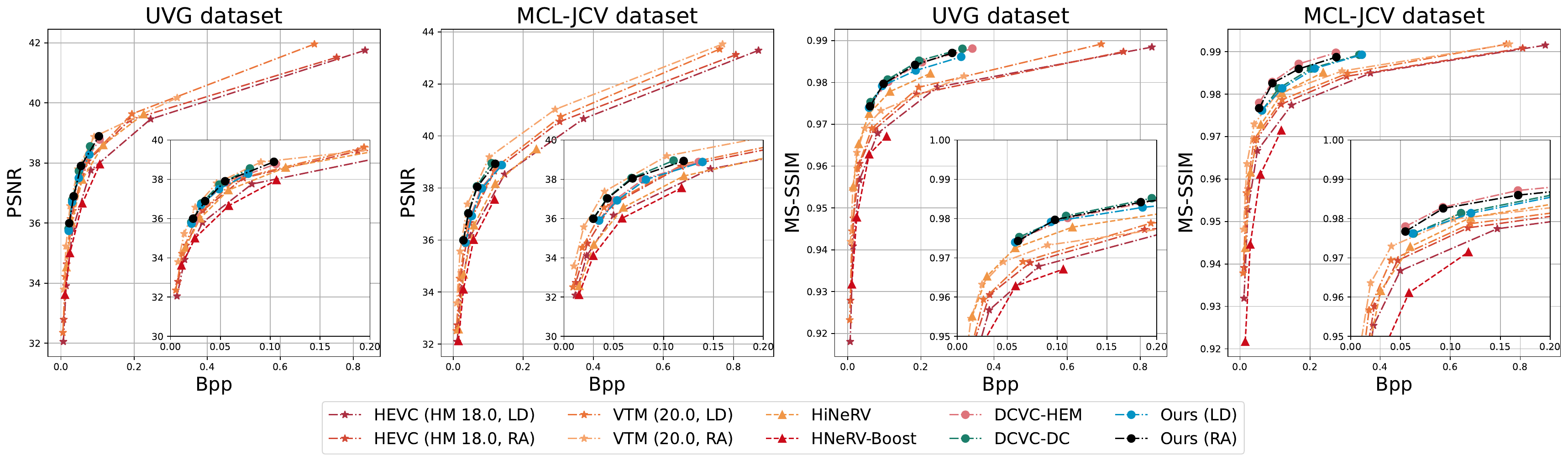}
    \caption{RD performance comparison on UVG and MCL-JCV dataset, where the two best performers for each type is plotted.}
    \label{fig:rd-plot}
\end{figure*}

\section{Experiments} 
\subsubsection{Datasets.} \label{sec:datasets} The proposed PNVC codec is pre-trained on the Vimeo-90k~\cite{xue2019video} dataset and evaluated on the UVG~\cite{mercat2020uvg} and MCL-JCV~\cite{wang2016mcl} databases that are 1080P. 

\subsubsection{Baselines.} The proposed model is compared with nine open-sourced state-of-the-art conventional and neural video compression baselines including (i) two conventional codecs - H.265/HEVC Test Model HM 18.0~\cite{hm} and H.266/VVC Test Model VTM 20.0 ~\cite{vtm}; (ii) three neural video codecs - VCT~\cite{mentzer2022vct}, DCVC-HEM~\cite{li2022hybrid}, and DCVC-DC~\cite{li2023neural}; (iii) four INR-based codecs - FFNeRV~\cite{lee2023ffnerv}, HNeRV-Boost~\cite{zhang2024boosting}, C3~\cite{kim2024c3} and HiNeRV~\cite{kwan2024hinerv}. 

\subsubsection{Test conditions.} Two coding configurations are employed: Low Delay (LD) and Random Access (RA), following the VTM common test conditions (CTC) \cite{vtm_ctc}. The LD configuration requires only one intra frame in each sequence (the first one), with subsequent P frames relying solely on previous frames for motion prediction (\textit{i.e.}, GOP=1) as defined in JVET CTC. In RA mode, each GOP consists of one intra frame and 31 B/P frames, allowing a maximum latency of 31 (\textit{i.e.}, GOP=32) with an IntraPeriod of 32. The RA configuration here uses the same hierarchical B frame structure as specified in the JVET CTC. These configurations do not apply to other INR-based benchmarks that encode the entire sequence or dataset. 

\subsubsection{Metrics.} \label{sec:metrics} For each test rate point of a video codec, we calculate the bitrate (in terms of bit/pixel, bpp) and their video quality in terms of PSNR and MS-SSIM in the RGB space. Based on these, we further calculate the Bj{\o}ntegaard Delta Rate (BD-rate)~\cite{bdrate} using HM (LD) as the anchor. We also measured the complexity figures of each tested codec, including the number of model parameters, the decoding complexity (MACs/pixel) and average encoding and decoding speeds (FPS).

\section{Results}

\begin{figure*}[htb!]
    \centering
    \includegraphics[width=0.972\textwidth]{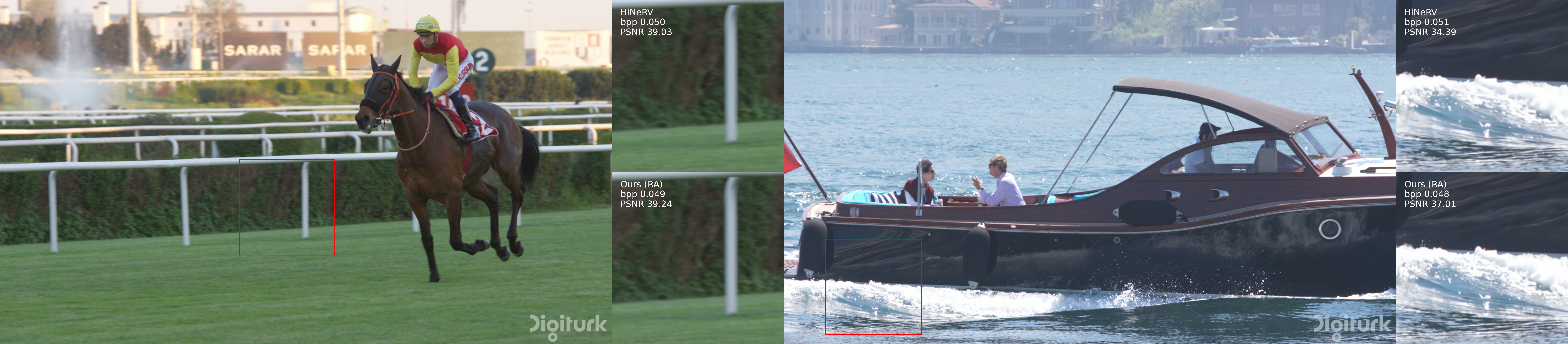}
    \caption{Visual quality comparison between HiNeRV and the proposed PNVC reconstructed content at similar bitrates.}
    \label{fig:visual-comparison}
\end{figure*}

\subsection{Quantitative results}
The BD-rate performance is summarized in Table~\ref{tab:bd-psnr}, where it can be seen that  PNVC (LD and RA modes) outperforms the anchor HM 18.0 (LD) by 34.35\% and 39.98\% in BD-rate in terms of PSNR on the UVG and MCL-JCV dataset, respectively. PNVC also significantly outperforms state-of-the-art INR-based codecs and is also competitive against the state-of-the-art conventional and neural video codecs. Furthermore, it should be noted that PNVC embodies latency constraints (31 frames for the RA mode and 0 for the LD mode, the same as HM and VTM), whereas other INR models do not. These demonstrate the strong rate-distortion performance of the proposed model. Among all codecs tested, the next-generation standard VTM (RA) remains to be the best performing baseline. 

\subsection{Qualitative results}
We present qualitative visual comparisons between frames reconstructed by HiNeRV (the best INR-based video codec) and the proposed PNVC in Figure~\ref{fig:visual-comparison}. The visual results show that our method is able to yield visually pleasing reconstructions, accurately capturing the fine texture details and fast/complex motions present in the frame. Please see \textit{Supplementary} for more thorough comparisons.

\subsection{Complexity performance}
The complexity figures of PNVC and other benchmark codecs, including encoding and decoding speeds (FPS), full model size (number of parameters) and decoding complexity (MACs/pixel), are summarized in Table~\ref{tab:bd-psnr}. These are measured by averaging over a GOP of 1080P videos, including a full roll-out of the entropy coding process, on a PC with an NVIDIA 3090 GPU and an Intel Core i7-12700 CPU. It can be observed that our model achieves a higher decoding speed, comparable to the conventional and INR baselines and considerably faster than all the neural video codecs. The PNVC decoder is also similar in size to INR models and much smaller than neural codecs. The well-roundness of our proposed model is illustrated by the Radar plot, as shown in Figure~\ref{fig:performance-complexity}, in which decoding speed, encoding latency, BD-rate in PSNR and MS-SSIM of the selected, top-performing baselines are visualized. It is evident that the proposed PNVC yields the most-balanced and also the best overall rate-distortion-complexity trade-off. It is also noted that the encoding speed of PNVC is still relatively low (so as other INR-based codecs), when compared to autoencoder-based neural compression models (and VTM), which limits real-time deployment (\textit{e.g.,} video conferencing). However, the benefits of Low Delay and Random Access compatibility, low decoding complexity, and competitive rate quality performance still offer great potential to support widely used streaming scenarios, which does not require real-time encoding ability.    

\subsection{Ablation studies}
To validate the contribution of each design component in the proposed PNVC model, we conducted ablations using the following model variants. The corresponding BD-rate results are summarized in Table~\ref{tab:ablation}.

\begin{table}[!t]
    \caption{Ablation study results in terms of BD-rate w.r.t the full PNVC codec. Positive values indicate coding loss. }
    \centering
    \label{tab:ablation}
     \resizebox{\columnwidth}{!}{\begin{tabular}{l|l|cc}
        \toprule
        Version & Ablation option & UVG & MCL-JCV \\
        \midrule
        (V1.1) & \xmark{} re-param. @ pre-train. & 6.45\% & 6.34\% \\
        (V1.2) & \xmark{} re-param. @ overfit. & 3.43\% & 3.25\% \\
        (V1.3) & \xmark{} re-param. @ fully-connected & 2.22\% & 2.36\% \\
        (V1.4) & \xmark{} re-param. @ token-mix  & 4.57\% & 4.49\% \\
        \midrule
        (V2.1) & \xmark{} adaptive QP (LD) & 1.21\% & 1.22\% \\
        (V2.2) & \xmark{} adaptive QP (RA) & 2.25\% & 2.23\% \\
        (V2.3) & \xmark{} overfitting (LD) & 7.95\% & 7.95\% \\
        (V2.4) & \xmark{} overfitting (RA) & 9.47\% & 9.34\% \\
        \midrule
        (V3.1) & \xmark{} uneven channel grouping & 1.76\% & 1.91\% \\
        (V3.2) & \xmark{} modulation \cmark{} concat. & 3.39\% & 3.57\% \\
        \midrule
        (V4.1) & \xmark{} scaled hierarchical griding & 2.39\% & 2.36\% \\
        (V4.2) & \xmark{} rescaling $\vect{\gamma}$ & 0.75\% & 0.76\% \\
        (V4.3) & \xmark{} shifting $\vect{\beta}$ & 0.89\% & 0.91\% \\
        \bottomrule
    \end{tabular}}
\end{table}

\subsubsection{Reparameterization.} We tested the effectiveness of reparameterization by removing it from pre-training (V1.1), performing online overfitting with the pre-trained model but disregarding the overparameterized basis (V1.2), removing the reparameterization for MLP (V1.3), and removing the degenerated token mixer component (V1.4).

\subsubsection{Optimization.} We also validated the adaptive quality control parameter $q_{\text{glob};t}$ in both the LD and RA coding configurations by removing them from the optimization process (V2.1 and V2.2, respectively). The coding performance of pre-training only without overfitting is ablated in (V2.3).

\subsubsection{Entropy model.} We adopted the original quadtree-chunking strategy instead of using uneven channel grouping (V3.1) and replaced temporal feature modulation with concatenation (V3.2) to verify the effectiveness of the improved entropy model.

\subsubsection{Decoder.} We respectively replaced the adaptive rescaled hierarchical encoding (V4.1) with the original hierarchical griding scheme (overfitted for the same number of steps), the rescaling feature $\{\vect{\gamma}^{l}_t\}^L_{l=1}$ (V4.2) or the shifting feature $\{\vect{\beta}^{l}_t\}^L_{l=1}$ (V4.3) to quantify their contributions to the overall performance gain. 

It is noted that all the thirteen tested variants perform worse than PNVC on both datasets, as shown in Table~\ref{tab:ablation}, demonstrating the effectiveness of each design component.

\section{Conclusion}
We propose PNVC, an INR-based video coding framework which 1) combines autoencoder-based compression methods with implicit, coordinate-based overfitting and 2) integrates both offline pretraining and online overfitting. PNVC supports Low Delay and Random Access coding modes and achieves competitive rate-distortion-complexity performance with fast decoding speed. Through innovations including reparameterized visual tuning, hierarchical quality parameters, modulation-based entropy modeling with asymmetric channel partition, and scale-adaptive hierarchical griding, PNVC outperforms VTM 20.0 (LD) and HiNeRV, matches DCVC-DC in compression efficiency, and maintains a low decoding complexity of $<$200k MACs/pixel and a decoding speed of $>$20 FPS @1080P. More importantly, PNVC supports flexible latency configurations (LD and RA), which improves the practicality of INR-based video codecs. Future studies must address could exploit meta-learning~\cite{lee2021meta} methods to speed up the encoding process.

\subsubsection{Acknowledgments.}
This work was supported by the UK Research and Innovation (UKRI) MyWorld Strength in Places Program (SIPF00006/1).

\bibliography{aaai25}

\begin{thebibliography}{59}
\providecommand{\natexlab}[1]{#1}

\bibitem[{Agustsson et~al.(2020)Agustsson, Minnen, Johnston, Balle, Hwang, and Toderici}]{agustsson2020scale}
Agustsson, E.; Minnen, D.; Johnston, N.; Balle, J.; Hwang, S.~J.; and Toderici, G. 2020.
\newblock Scale-space flow for end-to-end optimized video compression.
\newblock In \emph{Proceedings of the IEEE/CVF Conference on Computer Vision and Pattern Recognition}, 8503--8512.

\bibitem[{Bai et~al.(2023)Bai, Dong, Wang, and Yuan}]{bai2023ps}
Bai, Y.; Dong, C.; Wang, C.; and Yuan, C. 2023.
\newblock {PS-NeRV}: Patch-wise stylized neural representations for videos.
\newblock In \emph{IEEE International Conference on Image Processing}, 41--45. IEEE.

\bibitem[{Ball{\'e} et~al.(2018)Ball{\'e}, Minnen, Singh, Hwang, and Johnston}]{balle2018variational}
Ball{\'e}, J.; Minnen, D.; Singh, S.; Hwang, S.~J.; and Johnston, N. 2018.
\newblock Variational image compression with a scale hyperprior.
\newblock In \emph{International Conference on Learning Representations}.

\bibitem[{Bjontegaard(2001)}]{bdrate}
Bjontegaard, G. 2001.
\newblock Calculation of average PSNR differences between RD-curves.
\newblock \emph{ITU SG16 Doc. VCEG-M33}.

\bibitem[{Bossen et~al.(2023)Bossen, Boyce, Suehring, Li, and Seregin}]{vtm_ctc}
Bossen, F.; Boyce, J.; Suehring, K.; Li, X.; and Seregin, V. 2023.
\newblock {VTM Common Test Conditions and Software Reference Configurations for SDR Video}.
\newblock In \emph{the JVET meeting}, {JVET-T2010}.

\bibitem[{Bross et~al.(2021)Bross, Wang, Ye, Liu, Chen, Sullivan, and Ohm}]{bross2021overview}
Bross, B.; Wang, Y.-K.; Ye, Y.; Liu, S.; Chen, J.; Sullivan, G.~J.; and Ohm, J.-R. 2021.
\newblock Overview of the versatile video coding ({VVC}) standard and its applications.
\newblock \emph{IEEE Transactions on Circuits and Systems for Video Technology}, 31(10): 3736--3764.

\bibitem[{Browne, Ye, and Kim(2023)}]{vtm}
Browne, A.; Ye, Y.; and Kim, S.~H. 2023.
\newblock {Algorithm description for Versatile Video Coding and Test Model 19 (VTM 19)}.
\newblock In \emph{the JVET meeting}, {JVET-AC2002}. ITU-T and ISO/IEC.

\bibitem[{Chen et~al.(2023)Chen, Gwilliam, Lim, and Shrivastava}]{chen2023hnerv}
Chen, H.; Gwilliam, M.; Lim, S.-N.; and Shrivastava, A. 2023.
\newblock {HNeRV}: A hybrid neural representation for videos.
\newblock In \emph{Proceedings of the IEEE/CVF Conference on Computer Vision and Pattern Recognition}, 10270--10279.

\bibitem[{Chen et~al.(2021)Chen, He, Wang, Ren, Lim, and Shrivastava}]{chen2021nerv}
Chen, H.; He, B.; Wang, H.; Ren, Y.; Lim, S.~N.; and Shrivastava, A. 2021.
\newblock {NeRV}: Neural representations for videos.
\newblock \emph{Advances in Neural Information Processing Systems}, 34: 21557--21568.

\bibitem[{Ding et~al.(2021)Ding, Xia, Zhang, Chu, Han, and Ding}]{ding2021repmlp}
Ding, X.; Xia, C.; Zhang, X.; Chu, X.; Han, J.; and Ding, G. 2021.
\newblock Repmlp: Re-parameterizing convolutions into fully-connected layers for image recognition.
\newblock \emph{arXiv preprint arXiv:2105.01883}.

\bibitem[{Ding et~al.(2022)Ding, Zhang, Han, and Ding}]{ding2022scaling}
Ding, X.; Zhang, X.; Han, J.; and Ding, G. 2022.
\newblock Scaling up your kernels to 31x31: Revisiting large kernel design in cnns.
\newblock In \emph{{IEEE/CVF Conference on Computer Vision and Pattern Recognition}}, 11963--11975.

\bibitem[{Dupont et~al.(2021)Dupont, Golinski, Alizadeh, Teh, and Doucet}]{dupont2021coin}
Dupont, E.; Golinski, A.; Alizadeh, M.; Teh, Y.~W.; and Doucet, A. 2021.
\newblock {COIN}: Compression with Implicit Neural representations.
\newblock In \emph{ICLR Workshop on Neural Compression: From Information Theory to Applications}.

\bibitem[{Gomes, Azevedo, and Schroers(2023)}]{gomes2023video}
Gomes, C.; Azevedo, R.; and Schroers, C. 2023.
\newblock Video compression with entropy-constrained neural representations.
\newblock In \emph{Proceedings of the IEEE/CVF Conference on Computer Vision and Pattern Recognition}, 18497--18506.

\bibitem[{He et~al.(2023)He, Yang, Wang, Wu, Chen, Huang, Ren, Lim, and Shrivastava}]{he2023towards}
He, B.; Yang, X.; Wang, H.; Wu, Z.; Chen, H.; Huang, S.; Ren, Y.; Lim, S.-N.; and Shrivastava, A. 2023.
\newblock Towards scalable neural representation for diverse videos.
\newblock In \emph{Proceedings of the IEEE/CVF Conference on Computer Vision and Pattern Recognition}, 6132--6142.

\bibitem[{He et~al.(2022)He, Yang, Peng, Ma, Qin, and Wang}]{he2022elic}
He, D.; Yang, Z.; Peng, W.; Ma, R.; Qin, H.; and Wang, Y. 2022.
\newblock {ELIC}: Efficient learned image compression with unevenly grouped space-channel contextual adaptive coding.
\newblock In \emph{Proceedings of the IEEE/CVF Conference on Computer Vision and Pattern Recognition}, 5718--5727.

\bibitem[{Ho et~al.(2022)Ho, Chang, Chen, Gnutti, and Peng}]{ho2022canf}
Ho, Y.-H.; Chang, C.-P.; Chen, P.-Y.; Gnutti, A.; and Peng, W.-H. 2022.
\newblock {CANF-VC}: Conditional augmented normalizing flows for video compression.
\newblock In \emph{European Conference on Computer Vision}, 207--223. Springer.

\bibitem[{Hu, Lu, and Xu(2021)}]{hu2021fvc}
Hu, Z.; Lu, G.; and Xu, D. 2021.
\newblock {FVC}: A new framework towards deep video compression in feature space.
\newblock In \emph{Proceedings of the IEEE/CVF Conference on Computer Vision and Pattern Recognition}, 1502--1511.

\bibitem[{Hu and Xu(2023)}]{hu2023complexity}
Hu, Z.; and Xu, D. 2023.
\newblock {Complexity-guided slimmable decoder for efficient deep video compression}.
\newblock In \emph{{IEEE/CVF Conference on Computer Vision and Pattern Recognition}}, 14358--14367.

\bibitem[{Huang and Belongie(2017)}]{huang2017arbitrary}
Huang, X.; and Belongie, S. 2017.
\newblock Arbitrary style transfer in real-time with adaptive instance normalization.
\newblock In \emph{Proceedings of the IEEE International Conference on Computer Vision}, 1501--1510.

\bibitem[{Khani, Sivaraman, and Alizadeh(2021)}]{khani2021efficient}
Khani, M.; Sivaraman, V.; and Alizadeh, M. 2021.
\newblock Efficient video compression via content-adaptive super-resolution.
\newblock In \emph{Proceedings of the IEEE/CVF International Conference on Computer Vision}, 4521--4530.

\bibitem[{Kim et~al.(2024)Kim, Bauer, Theis, Schwarz, and Dupont}]{kim2024c3}
Kim, H.; Bauer, M.; Theis, L.; Schwarz, J.~R.; and Dupont, E. 2024.
\newblock C3: High-performance and low-complexity neural compression from a single image or video.
\newblock In \emph{Proceedings of the IEEE/CVF Conference on Computer Vision and Pattern Recognition}, 9347--9358.

\bibitem[{Kwan et~al.(2024{\natexlab{a}})Kwan, Gao, Zhang, Gower, and Bull}]{kwan2024hinerv}
Kwan, H.~M.; Gao, G.; Zhang, F.; Gower, A.; and Bull, D. 2024{\natexlab{a}}.
\newblock HiNeRV: Video Compression with Hierarchical Encoding-based Neural Representation.
\newblock \emph{{Advances in Neural Information Processing Systems}}, 36: 72692--72704.

\bibitem[{Kwan et~al.(2024{\natexlab{b}})Kwan, Zhang, Gower, and Bull}]{kwan2024immersive}
Kwan, H.~M.; Zhang, F.; Gower, A.; and Bull, D. 2024{\natexlab{b}}.
\newblock Immersive Video Compression using Implicit Neural Representations.
\newblock In \emph{Picture Coding Symposium}.

\bibitem[{Lee et~al.(2021)Lee, Tack, Lee, and Shin}]{lee2021meta}
Lee, J.; Tack, J.; Lee, N.; and Shin, J. 2021.
\newblock Meta-learning sparse implicit neural representations.
\newblock \emph{{Advances in Neural Information Processing Systems}}, 34: 11769--11780.

\bibitem[{Lee et~al.(2023)Lee, Rho, Ko, and Park}]{lee2023ffnerv}
Lee, J.~C.; Rho, D.; Ko, J.~H.; and Park, E. 2023.
\newblock {FFNeRV}: Flow-guided frame-wise neural representations for videos.
\newblock In \emph{Proceedings of the 31st ACM International Conference on Multimedia}, 7859--7870.

\bibitem[{Lee et~al.(2024)Lee, Rho, Nam, Ko, and Park}]{lee2024coordinate}
Lee, J.~C.; Rho, D.; Nam, S.; Ko, J.~H.; and Park, E. 2024.
\newblock Coordinate-aware modulation for neural fields.
\newblock In \emph{The Twelfth International Conference on Learning Representations}.

\bibitem[{Leguay et~al.(2024)Leguay, Ladune, Philippe, and D{\'e}forges}]{leguay2024cool}
Leguay, T.; Ladune, T.; Philippe, P.; and D{\'e}forges, O. 2024.
\newblock Cool-chic video: Learned video coding with 800 parameters.
\newblock \emph{arXiv preprint arXiv:2402.03179}.

\bibitem[{Li, Li, and Lu(2021)}]{li2021deep}
Li, J.; Li, B.; and Lu, Y. 2021.
\newblock {Deep contextual video compression}.
\newblock \emph{{Advances in Neural Information Processing Systems}}, 34: 18114--18125.

\bibitem[{Li, Li, and Lu(2022)}]{li2022hybrid}
Li, J.; Li, B.; and Lu, Y. 2022.
\newblock Hybrid spatial-temporal entropy modelling for neural video compression.
\newblock In \emph{Proceedings of the 30th ACM International Conference on Multimedia}, 1503--1511.

\bibitem[{Li, Li, and Lu(2023)}]{li2023neural}
Li, J.; Li, B.; and Lu, Y. 2023.
\newblock Neural Video Compression with Diverse Contexts.
\newblock In \emph{{IEEE/CVF Conference on Computer Vision and Pattern Recognition}}, 22616--22626.

\bibitem[{Li, Li, and Lu(2024)}]{li2024neural}
Li, J.; Li, B.; and Lu, Y. 2024.
\newblock Neural video compression with feature modulation.
\newblock In \emph{Proceedings of the IEEE/CVF Conference on Computer Vision and Pattern Recognition}, 26099--26108.

\bibitem[{Li et~al.(2022)Li, Wang, Pi, Xu, Mei, and Liu}]{li2022nerv}
Li, Z.; Wang, M.; Pi, H.; Xu, K.; Mei, J.; and Liu, Y. 2022.
\newblock {E-NeRV}: Expedite neural video representation with disentangled spatial-temporal context.
\newblock In \emph{European Conference on Computer Vision}, 267--284. Springer.

\bibitem[{Lin et~al.(2024)Lin, Lyu, Liu, Tang, Liang, Song, and Chang}]{lin2024mlp}
Lin, S.; Lyu, P.; Liu, D.; Tang, T.; Liang, X.; Song, A.; and Chang, X. 2024.
\newblock MLP Can Be A Good Transformer Learner.
\newblock In \emph{{IEEE/CVF Conference on Computer Vision and Pattern Recognition}}, 19489--19498.

\bibitem[{Lu et~al.(2020)Lu, Cai, Zhang, Chen, Ouyang, Xu, and Gao}]{lu2020content}
Lu, G.; Cai, C.; Zhang, X.; Chen, L.; Ouyang, W.; Xu, D.; and Gao, Z. 2020.
\newblock Content adaptive and error propagation aware deep video compression.
\newblock In \emph{European Conference on Computer Vision}, 456--472.

\bibitem[{Lu et~al.(2019)Lu, Ouyang, Xu, Zhang, Cai, and Gao}]{lu2019dvc}
Lu, G.; Ouyang, W.; Xu, D.; Zhang, X.; Cai, C.; and Gao, Z. 2019.
\newblock {DVC:} An End-To-End Deep Video Compression Framework.
\newblock In \emph{{IEEE/CVF Conference on Computer Vision and Pattern Recognition}}, 11006--11015.

\bibitem[{Lu et~al.(2024)Lu, Duan, Zhu, and Ma}]{lu2024deep}
Lu, M.; Duan, Z.; Zhu, F.; and Ma, Z. 2024.
\newblock Deep Hierarchical Video Compression.
\newblock In \emph{Proceedings of the AAAI Conference on Artificial Intelligence}, 8859--8867.

\bibitem[{Ma, Zhang, and Bull(2020)}]{ma2020mfrnet}
Ma, D.; Zhang, F.; and Bull, D.~R. 2020.
\newblock MFRNet: a new CNN architecture for post-processing and in-loop filtering.
\newblock \emph{IEEE Journal of Selected Topics in Signal Processing}, 15(2): 378--387.

\bibitem[{Mentzer et~al.(2022)Mentzer, Toderici, Minnen, Caelles, Hwang, Lucic, and Agustsson}]{mentzer2022vct}
Mentzer, F.; Toderici, G.~D.; Minnen, D.; Caelles, S.; Hwang, S.~J.; Lucic, M.; and Agustsson, E. 2022.
\newblock VCT: A Video Compression Transformer.
\newblock \emph{Advances in Neural Information Processing Systems}, 35: 13091--13103.

\bibitem[{Mercat, Viitanen, and Vanne(2020)}]{mercat2020uvg}
Mercat, A.; Viitanen, M.; and Vanne, J. 2020.
\newblock UVG dataset: 50/120fps 4K sequences for video codec analysis and development.
\newblock In \emph{Proceedings of the 11th ACM Multimedia Systems Conference}, 297--302.

\bibitem[{Minnen and Singh(2020)}]{minnen2020channel}
Minnen, D.; and Singh, S. 2020.
\newblock Channel-wise autoregressive entropy models for learned image compression.
\newblock In \emph{2020 IEEE International Conference on Image Processing}, 3339--3343. IEEE.

\bibitem[{Peng et~al.(2024)Peng, Gao, Sun, Zhang, and Bull}]{peng2023accelerating}
Peng, T.; Gao, G.; Sun, H.; Zhang, F.; and Bull, D. 2024.
\newblock Accelerating Learnt Video Codecs with Gradient Decay and Layer-Wise Distillation.
\newblock In \emph{Picture Coding Symposium}.

\bibitem[{Ranjan and Black(2017)}]{ranjan2017optical}
Ranjan, A.; and Black, M.~J. 2017.
\newblock Optical flow estimation using a spatial pyramid network.
\newblock In \emph{Proceedings of the IEEE Conference on Computer Vision and Pattern Recognition}, 4161--4170.

\bibitem[{Ruan et~al.(2024)Ruan, Shao, Yang, Zhao, and Niyato}]{ruan2024point}
Ruan, H.; Shao, Y.; Yang, Q.; Zhao, L.; and Niyato, D. 2024.
\newblock Point Cloud Compression with Implicit Neural Representations: A Unified Framework.
\newblock \emph{arXiv preprint arXiv:2405.11493}.

\bibitem[{Sharman, Sjoberg, and Sullivan(2022)}]{hm}
Sharman, C. R.~K.; Sjoberg, R.; and Sullivan, G. 2022.
\newblock {High Efficiency Video Coding (HEVC) Test Model 16 (HM 16) Improved Encoder Description Update 16}.
\newblock In \emph{the JVET meeting}, {JVET-Y1002}. ITU-T and ISO/IEC.

\bibitem[{Shi, Zhou, and Gu(2024)}]{shi2024improved}
Shi, K.; Zhou, X.; and Gu, S. 2024.
\newblock Improved Implicity Neural Representation with Fourier Bases Reparameterized Training.
\newblock \emph{arXiv preprint arXiv:2401.07402}.

\bibitem[{Sitzmann et~al.(2020)Sitzmann, Martel, Bergman, Lindell, and Wetzstein}]{sitzmann2020implicit}
Sitzmann, V.; Martel, J.; Bergman, A.; Lindell, D.; and Wetzstein, G. 2020.
\newblock Implicit neural representations with periodic activation functions.
\newblock \emph{Advances in neural information processing systems}, 33: 7462--7473.

\bibitem[{Str{\"u}mpler et~al.(2022)Str{\"u}mpler, Postels, Yang, Gool, and Tombari}]{strumpler2022implicit}
Str{\"u}mpler, Y.; Postels, J.; Yang, R.; Gool, L.~V.; and Tombari, F. 2022.
\newblock Implicit neural representations for image compression.
\newblock In \emph{European Conference on Computer Vision}, 74--91.

\bibitem[{Sullivan et~al.(2012)Sullivan, Ohm, Han, and Wiegand}]{sullivan2012overview}
Sullivan, G.~J.; Ohm, J.; Han, W.; and Wiegand, T. 2012.
\newblock {Overview of the High Efficiency Video Coding (HEVC) Standard}.
\newblock \emph{{IEEE} Transactions on Circuits and Systems for Video Technology}, 22(12): 1649--1668.

\bibitem[{van Rozendaal, Huijben, and Cohen(2021)}]{van2021overfitting}
van Rozendaal, T.; Huijben, I.; and Cohen, T.~S. 2021.
\newblock Overfitting for fun and profit: Instance-adaptive data compression.
\newblock In \emph{International Conference on Learning Representations}.

\bibitem[{Wang et~al.(2023)Wang, Li, Li, and Lu}]{wang2023EVC}
Wang, G.-H.; Li, J.; Li, B.; and Lu, Y. 2023.
\newblock {EVC: Towards Real-Time Neural Image Compression with Mask Decay}.
\newblock In \emph{International Conference on Learning Representations}.

\bibitem[{Wang et~al.(2016)Wang, Gan, Hu, Lin, Jin, Song, Wang, Katsavounidis, Aaron, and Kuo}]{wang2016mcl}
Wang, H.; Gan, W.; Hu, S.; Lin, J.~Y.; Jin, L.; Song, L.; Wang, P.; Katsavounidis, I.; Aaron, A.; and Kuo, C.-C.~J. 2016.
\newblock MCL-JCV: a JND-based H. 264/AVC video quality assessment dataset.
\newblock In \emph{IEEE international conference on image processing}, 1509--1513.

\bibitem[{Wiegand et~al.(2003)Wiegand, Sullivan, Bj{\o}ntegaard, and Luthra}]{wiegand2003overview}
Wiegand, T.; Sullivan, G.~J.; Bj{\o}ntegaard, G.; and Luthra, A. 2003.
\newblock {Overview of the H.264/AVC Video Coding Standard}.
\newblock \emph{{IEEE} Transactions on Circuits and Systems for Video Technology}, 13(7): 560--576.

\bibitem[{Xiang, Tian, and Zhang(2022)}]{xiang2022mimt}
Xiang, J.; Tian, K.; and Zhang, J. 2022.
\newblock Mimt: Masked image modeling transformer for video compression.
\newblock In \emph{International Conference on Learning Representations}.

\bibitem[{Xue et~al.(2019)Xue, Chen, Wu, Wei, and Freeman}]{xue2019video}
Xue, T.; Chen, B.; Wu, J.; Wei, D.; and Freeman, W.~T. 2019.
\newblock Video enhancement with task-oriented flow.
\newblock \emph{International Journal of Computer Vision}, 127: 1106--1125.

\bibitem[{Yan et~al.(2018)Yan, Liu, Li, Li, Li, and Wu}]{yan2018convolutional}
Yan, N.; Liu, D.; Li, H.; Li, B.; Li, L.; and Wu, F. 2018.
\newblock Convolutional neural network-based fractional-pixel motion compensation.
\newblock \emph{IEEE Transactions on Circuits and Systems for Video Technology}, 29(3): 840--853.

\bibitem[{Yang, Oh, and Park(2024)}]{yang2024parameter}
Yang, H.; Oh, S.; and Park, E. 2024.
\newblock Parameter-Efficient Instance-Adaptive Neural Video Compression.
\newblock \emph{arXiv preprint arXiv:2405.08530}.

\bibitem[{Zhang et~al.(2021{\natexlab{a}})Zhang, Ma, Feng, and Bull}]{zhang2021video}
Zhang, F.; Ma, D.; Feng, C.; and Bull, D.~R. 2021{\natexlab{a}}.
\newblock Video compression with CNN-based postprocessing.
\newblock \emph{IEEE MultiMedia}, 28(4): 74--83.

\bibitem[{Zhang et~al.(2024)Zhang, Yang, He, Ge, Xu, Wang, Qin, and Zhang}]{zhang2024boosting}
Zhang, X.; Yang, R.; He, D.; Ge, X.; Xu, T.; Wang, Y.; Qin, H.; and Zhang, J. 2024.
\newblock Boosting Neural Representations for Videos with a Conditional Decoder.
\newblock In \emph{Proceedings of the IEEE/CVF Conference on Computer Vision and Pattern Recognition}, 2556--2566.

\bibitem[{Zhang et~al.(2021{\natexlab{b}})Zhang, van Rozendaal, Brehmer, Nagel, and Cohen}]{zhang2021implicit}
Zhang, Y.; van Rozendaal, T.; Brehmer, J.; Nagel, M.; and Cohen, T. 2021{\natexlab{b}}.
\newblock Implicit Neural Video Compression.
\newblock In \emph{ICLR Workshop on Deep Generative Models for Highly Structured Data}.

\end{thebibliography}

\end{document}